\documentclass[journal]{IEEEtran}

\usepackage{float}
\usepackage{placeins}
\usepackage{cite}
\usepackage[table,xcdraw]{xcolor}

\usepackage[ruled,vlined]{algorithm2e}

\usepackage[noend]{algpseudocode}
\usepackage[utf8]{inputenc} % allow utf-8 input
\usepackage[T1]{fontenc}    % use 8-bit T1 fonts
\usepackage{hyperref}       % hyperlinks
\usepackage{url}            % simple URL typesetting
\usepackage{booktabs}       % professional-quality tables
\usepackage{amsfonts}       % blackboard math symbols
\usepackage{nicefrac}       % compact symbols for 1/2, etc.
\usepackage{microtype}      % microtypography
\usepackage{lipsum}
\usepackage{amsmath}
\newcommand{\minus}{$-$}
\usepackage{graphicx}
\usepackage{amsmath}
\graphicspath{ {./images/} }

\title{Adversarial Attacks on Multivariate Time Series}

\author{Samuel~Harford, %~\IEEEmembership{Graduate Student Member,~IEEE,}
        Fazle~Karim, %~\IEEEmembership{Graduate Student Member,~IEEE,}
        and~Houshang~Darabi%, ~\IEEEmembership{Senior Member,~IEEE}
        
\thanks{S. Harford, F. Karim and H. Darabi are with the Department of Mechanical and
Industrial Engineering, University of Illinois at Chicago, 842 West Taylor
Street, Chicago, IL 60607, United States. H. Darabi is the corresponding
author. E-mail: {sharfo2, karim1, hdarabi}@uic.edu}% <-this % stops a space
\thanks{Manuscript received MONTH XX, 2020; revised MONTH XX, 2020}}

\begin{document}
\maketitle
\begin{abstract}
Classification models for the multivariate time series have gained significant importance in the research community, but not much research has been done on generating adversarial samples for these models. Such samples of adversaries could become a security concern. In this paper, we propose transforming the existing adversarial transformation network (ATN) on a distilled model to attack various multivariate time series classification models. The proposed attack on the classification model utilizes a distilled model as a surrogate that mimics the behavior of the attacked classical multivariate time series classification models. The proposed methodology is tested onto 1-Nearest Neighbor Dynamic Time Warping (1-NN DTW) and a Fully Convolutional Network (FCN), all of which are trained on 18 University of East Anglia (UEA) and University of California Riverside (UCR) datasets. We show both models were susceptible to attacks on all 18 datasets. To the best of our knowledge, adversarial attacks have only been conducted in the domain of univariate time series and have not been conducted on multivariate time series. such an attack on time series classification models has never been done before. Additionally, we recommend future researchers that develop time series classification models to incorporating adversarial data samples into their training data sets to improve resilience on adversarial samples and to consider model robustness as an evaluative metric.

\end{abstract}

% keywords can be removed

\section{Introduction}
\hspace*{5mm}The past decade has seen numerous areas of research and society impacted by machine learning and deep learning \cite{lecun2015deep}. These areas include medical imaging \cite{lo1995artificial}, speech-recognition \cite{mitra2014medium}, and manufacturing systems \cite{doltsinis2019machine}. With the rise of smart sensors, vast scale developments in data collection and storage, ease of data analytics and predictive modeling, multivariate time series data recieved from collections of sensors can be analyzed to identify regular patterns that can be interpreted and exploited. Many researchers have been interested in the classification of both univariate \cite{sirisambhand2019dimensionality, sharabiani2017efficient, sharabiani2018asymptotic, xi2006fast} and multivariate time series \cite{pei2017multivariate, schafer2017multivariate, karim2019multivariate}. Time series classification models are used in healthcare, where multiple lead ECG data are used to determine diagnose cardiac ischemia, in gesture recognition, where posture-level data is used to classify human actions, and in manufacturing, where sensor data is used to identify product detects. The combination of multi-channel sensor data that tracks resources and safety systems, along with real-time analytics, creates the possibility of automated responses to undesired operational activities. An effective time series classification model can capture and generalize patterns of time series signals, so it can classify unseen data. Similarly, classification models in the field of computer vision take advantage of the underlying spatial structure in images. However, studies have shown that computer vision models incorrectly classify images that seem obvious to the human eye, this is referred to as an adversarial attack \cite{goodfellow2014explaining}. Complex models can be tricked to incorrectly classify data from a wide array of fields using several types of adversarial attacks. This is a serious security issue in machine learning, especially Deep Neural Networks (DNN), which is widely used for vision-based tasks where adding minor disruptions or carefully crafted noise to an input image may mislead the image classification algorithm to make inaccurate predictions with a high degree of confidence \cite{mopuri2018generalizable, reddy2018nag}. Although DNNs are state-of-the-art models across several fields for a number of classification tasks, including time series classification \cite{karim2019multivariate, karim2017lstm, hashida2019multi}, these vulnerabilities have a harmful impact on real-world applicability in domains where secure and reliable predictions are of paramount importance \cite{miyato2018virtual}. Compounding the severity of this issue, Papernot et al's work has shown that it is easy to transfer adversarial attacks on a particular classifier of computer vision to other similar classifiers \cite{papernot2016transferability}. The focus of attacks has only recently been shifted to time series classification models based on deep neural networks and traditional models \cite{oregi2018adversarial}.\\
\hspace*{5mm}Many adversarial sample creation strategies have been suggested to trick various DNN (state-of-the-art computer vision models) image classification models. Most of these techniques are targeted at the DNNs gradient information that makes them vulnerable to these attacks \cite{akhtar2018threat, madry2017towards, tramer2017ensemble}. Research into generating adversarial sample for time series classification models has been limited to the univariate time series \cite{karim2019adversarial}. In speech recognition activities that translate text-to-speech, there is one major security issue. Carlini and Wagner \cite{carlini2018audio} demonstrate how it is possible to attack text-to-speech classifiers. We also provide multiple audio clips where the voice is not correctly identified by a text-to-speech classifier, DeepSpeech. Certain security concerns may arise in healthcare systems that use time series classification algorithms, where it can be fooled into misdiagnosing patients that may influence their disease diagnosis. Algorithms used to detect and monitor seismic activity in time series classification can be manipulated to create fear and hysteria in our society. Wearables that use time series data to classify activity of the wearer can be fooled into convincing the users they are doing other actions. Most of the current state-of-the-art multivariate time series classification algorithms are traditional approaches, such as 1-Nearest Neighbor Dynamic Time Warping (1-NN DTW) \cite{seto2015multivariate}, WEASEL+MUSE \cite{schafer2017multivariate} and Hidden-Unit Logistic Model \cite{pei2017multivariate}. However, due to their simplicity and effectiveness, DNNs are quickly becoming excellent time series classifiers. Traditional time series classification models are more difficult to attack as it can be considered a black-box model with an internal computation that is not differentiable. As such, it is impossible to exploit any gradient knowledge. However, as their gradient knowledge can be easily exploited, DNN models are more vulnerable to white-box attacks. A white-box attack is where the opponent has "given access to all elements of the training procedure"\cite{tramer2017ensemble}, including the training data set, the training algorithm, the model's parameters and weights, and the model architecture itself. In comparison, a black-box attack only has access to the training process and architecture of the target models\cite{tramer2017ensemble}.\\
\hspace*{5mm}This study proposes a proxy attack strategy on a target classifier via a student model, trained using standard model distillation techniques to mimic the behavior of the target multivariate time series classification models. The student network is the neural network distilled from another time series classification model, called the teacher model, that learns to approximate the output of the teacher model. Once the student model has been trained, our adversarial transformation network (ATN) is then trained to attack this student model. Our methodogolies are applied onto 1-NN DTW and Fully Convolutional Network (FCN) that are trained on 18 multivariate time series bench marks from the University of East Anglia (UEA) and the University of California, Riverside (UCR) \cite{DBLP:journals/corr/abs-1811-00075}. To the best of our knowledge, the result of such an attack on multivariate time series classification models has never been studied before. Finally, we recommend researchers that develop time series classification models to consider model robustness as an evaluative metric and incorporate adversarial data samples into their training data sets in order to further improve resilience to adversarial attacks.\\
\hspace*{5mm}The remainder of this paper is structured as follows: Section \ref{section:2} provides a background on the utilized multivariate time series classification models and information on adversarial crafting techniques used on computer vision problems. Section \ref{section:3} details our proposed methodologies. Section \ref{section:4} presents and explains the results of our proposed methodologies on a set of multivariate time series classification models. Section \ref{section:5} concludes the paper and proposes future work.

\section{Background and Related Works}
\label{section:2}

\subsection{Time Series Classifiers}
\subsubsection{Multivariate 1-Nearest Neighbor Dymanic Time Warping} The equations below for 1-Nearest Neighbor Dynamic Time Warping (1-NN DTW) are derived \cite{xi2006fast, kate2016using}. Dynamic Time Warping is a distance metric used to non-linearly align two time series. DTW outputs a matrix of the distance path and the shortest distance between series \cite{keogh2005exact}. For two time series $X = x_1, x_2, \dots, x_n$ and $Y = y_1, y_2, \dots, y_m$ of lengths $n$ and $m$ respectively, a DTW matrix can be calculated of size $n$ x $m$. Each cell of the DTW matrix represents an alignment between two points of the corresponding time series. Matrix calculations must follow the following conditions:
\textbf{Boundary Condition:} The paths should start from the beginning of each
time series $(x_1, y_1)$ and finish at the last point of each time series $(x_n, y_m)$.
\textbf{Continuity Condition:} The paths should have no jumps in steps; the points to consider for distance at the $(i, j)$ point are $(i \minus 1, j)$, $(i, j \minus 1)$, and $(i \minus 1, j \minus 1)$.
\textbf{Monotonicity Condition:} The warping paths can only go forward in time. $P$ is defined as a continuous path of cell in the matrix from $(x_1, y_1)$ to $(x_n, y_m)$. The $s^{th}$ element of $P$ is defined as $p_s = d(i, j)_s$; where $d(i, j) = (x_i \minus y_j)^2$, S is the length of $P = p_1, p_2, \dots, p_S$. The distance for DTW is equal to $\sqrt{D(n, m)}$, where the distance of each represented cell is found in Equation \eqref{eq:1}.

\begin{equation}
\label{eq:1}
D(i, j) = (x_i - y_j)^2 + min[D(i - 1, j - 1), D(i - 1, j), D(i, j - 1)]
\end{equation}

Initiated by the following conditions:
\begin{equation}
\label{eq:2}
D(1, 1) = 0; D(1, 2 \dots m) = \infty; D(2 \dots n, 1) = \infty
\end{equation}

\hspace*{5mm}In the case of multivariate time series, the DTW distance matrix is calculated between each channel of two multivariate time series. The summation of the DTW distances for each channel is then used as the similarity metric between the two multivariate time series. 

\subsubsection{Multi-Fully Convolutional Network}
The Multivariate Fully Convolutional Network (Multi-FCN) is one of the first deep learning networks used for the task of time series classification \cite{wang2017time}. The Multi-FCN architecture is an extension of the original FCN model that takes a univariate time series as input. Multi-FCN consists of 3 2D-Convolutional layers, with convolution kernels of size 8, 5 and 3 respectively, that emit 128, 256 and 128 filters respectively. Each convolution layer is followed by a batch normalization layer \cite{ioffe2015batch} with a ReLU activation layer. A global average pooling layer is applied after the final ReLU activation layer. The pooling layer is then passed to a softmax layer to determine the class probability vector.

\subsection{Adversarial Transformation Network}
\hspace*{5mm}Multiple different approaches for generating adversarial samples have been proposed to attack neural networks. These methods have focused on the task of generating adversaries for computer vision tasks. Most of these methods use either the gradient with respect to the image pixels of these neural networks or explicitly solving an optimization on the image pixel. Baluja and Fischer \cite{baluja2017adversarial} propose Adversarial Transformation Networks (ATNs) to efficiently generate an adversarial sample to attack networks by first using a self-supervised method to train a feed-forward neural network. Given the original input sample, ATNs modify the classifier outputs slightly to match the adversarial target. ATNs can be parametrize as a neural network $g_f(x) : x \rightarrow \hat{x}$, where $f$ is the target model (a time series classifier) which outputs either a class probability vector or a sparse class label, and $\hat{x} \sim x$, but $argmax$ $f(x) \neq$ $argmax$ $f(\hat{x})$. To find $g_f$, minimize the following loss function :

\begin{equation}
\label{eq:3}
L = \beta \star  L_x(g_f(x_i), x_i) + L_y(f(g_f (x_i)), f(x_i))
\end{equation}

where $L_x$ is a loss function on the input space (e.g. $L_2$ loss function), $L_y$ is the specially constructed loss function on the output space of $f$ to avoid learning the identity function, $x_i$ is the $i^{th}$ sample in the dataset and $\beta$ is the weighing term between the two loss functions. It is necessary to carefully select the loss function $L_y$ on the output space to successfully avoid learning the identity function. Baluja and Fischer \cite{baluja2017adversarial} define the loss function $L_y$ as $L_y(y', y) = L_2(y', r(y, t))$, where $y = f(x)$, $y' = f(g_f (x))$, and $r(·)$ is a reranking function that modifies $y$ such that $y_k < yt$, $\forall k \neq t$. This reranking function $r(y, t)$ can either be a simple one hot encoding function or can be formulated to take advantage of the already present $y$ to encourage better reconstruction. The reranking function proposed by Baluja and Fischer \cite{baluja2017adversarial} can be formulated as:

\begin{equation}
\label{eq:4}
r \alpha(y, t) = 
            norm\Biggl(
                    \biggl\{
                        \begin{aligned}
                        &\alpha \star max(y) &\text{ if } k=t,\\[1ex]
                        &y_k &\text{ } otherwise
                        \end{aligned}
                    \biggr\}_{\!\!k \in y}
                \Biggr)\\
\end{equation}

where $\alpha > 1$ is an additional hyperparameter which defines how much larger $y_t$ should be than the current max classification and norm is a normalizing function that rescales its input to be a valid probability distribution

\subsection{Transferability Property}
\hspace*{5mm}Papernot et al. \cite{papernot2016transferability} propose a black-box attack by training a local substitute network, $s$, to replicate or approximate the target deep neural network (DNN) model, $f$. The local substitute model is trained using synthetically generated samples and the output of these samples are labels from $f$. The local substitute network is than used to generate adversarial samples that are misclassifications. Generating adversarial samples for $s$ is much easier then generating adversaries from $f$, as its full knowledge/parameters are available, making it susceptible to various attacks. The key criteria to successfully generate adversarial samples of $f$ is the transferability property, where adversarial samples that misclassify $s$ will also misclassify $f$.

\subsection{Knowledge Distillation}
\hspace*{5mm}Knowledge distillation, first proposed by Bucila et al. \cite{bucilua2006model}, is a model compression technique where a small student model, $s$, is trained to mimic a pretrained teacher model, $f$. This process is also known as the model distillation training. The knowledge that is distilled from $f$ to $s$ is done by minimizing a loss function, where the objective of $s$ is to imitate the probability class vector output by the model $f$. Hinton et al. \cite{hinton2015distilling} note that there are several instances where the probability distribution is skewed such that the correct class probability would have a probability close to 1 and the remaining classes would have a probability closer to 0. For this reason, Hinton et al. \cite{hinton2015distilling} recommend computing the probabilities $q_i$ from the prenormalized logits $z_i$, such that:

\begin{equation}
\label{eq:5}
    q_i = \sigma(z; T) = exp(z_i/T)/sum_{j}exp(z_j/T)
\end{equation}

where $T$ is a temperature factor normally set to 1. Higher values of $T$ produce softer probability distributions over classes. The loss that is minimized is the model distillation loss, further explained in Section \ref{section:3c}.

\section{Methodology}
\label{section:3}

\subsection{Gradient Adversarial Transformation Network}
\label{section:3a}
\hspace*{5mm}This work studies black-box and white-box attacks on multivariate time series. Both attacks use methodologies expanded from ATNs \cite{baluja2017adversarial}. These ATNs are generative neural networks that take a multivariate time series $x$ as an input and outputs an adversarial sample $\hat{x}$.\\
\hspace*{5mm}An Adversarial Transformation Network can be parametrize as a neural network $g_f(x) : x \rightarrow \hat{x}$, where $f$ is the model to be attacked. The ATN is further adjusted with the gradient of the input sample $x$ with respect to the softmax scaled logits of the target class output by the attacked classifier. This adjustment results in the Gradient Adversarial Transformation Network (GATN) as a neural network $g_f(x, \tilde{x}) : (x, \tilde{x}) \rightarrow \hat{x}$, where:

\begin{equation}
\label{eq:6}
    \tilde{x} = \dfrac{\partial{x}}{\partial{f_t}}
\end{equation}

such that $x \in \mathbb{R}^T$ is an input multivariate time series of maximum length $T$, $f_t$ represents the probability of the input series being classified as class $t$. Given the input gradient $\tilde{x}$, the GANT can construct better adversarial samples to affect the targeted model and reduce the perturbation added to the sample. For this reason, the GANT model is used for all our attacks. \\
\hspace*{5mm} This study focuses on attacking 1-NN DTW and Multi-FCN time series classifiers. The 1-NN DTW classifier is non-differentiable, which creates a problem for the GATN model. A solution to overcoming the non-differentiability issue is discussed in Section \ref{section:3d}, by training a student network $s$ to approximate the output of the non-differentiable time series classifier $f$.

\subsection{Black-box and White-box Restrictions}
\label{section:3b}
\hspace*{5mm} The formulation presented in Section \ref{section:3a}. is satisfactory for white-box attacks, where the attacked model $f$ or the student model $s$ is known. For black-box attacks, we do not have access to the time series classifier or the dataset for model training. A further restriction for black-box attacks is to utilize only the outputted predicted label, and not the probabilistic class vector obtained from either a softmax layer or probabilistic approximations for classical model outputs.\\
\hspace*{5mm}For each dataset $D$, the training data is split into two halves. The GATN is trained on one half of the training data $D_{train}$. The remaining training data $D_{eval}$ is used to perform evaluations. In addition to $D_{eval}$, the unseen test set $D_{test}$ is used for evaluation. The available dataset $D$ is not the dataset that the attacked model $f$ was trained on. The available training set of the attacked model is never utilized to train or evaluate the GATN model. To satisfy these constraints, the available dataset $D$ is defined as the test set of the multivariate time series classification task. This test set is not used to train any attacked model $f$, therefore it can be used as an unseen dataset. The test dataset is then split into two halves with equivalent class balance. When evaluating black-box attacks, the available dataset is treated as if it were unlabeled. Due to this restriction, the predicted label from the attacked model $f$ is utilized to label the dataset prior to the attacks. This restriction adds realism to the training of GATNs, as it is difficult to obtain or create labeled datasets for time series tasks compared to computer vision tasks. 

\subsection{Training Methodology}
\label{section:3c}
\hspace*{5mm}When training models ATN and GATN, the selected reranking function $r(\cdot)$ strongly affects the loss formulation on the prediction space ($L y$). When we opt for the one hot encoding of the target class, we lose the ability to keep class ordering and the ability to adjust the ranking weight ($\alpha$) to get less skewed adversaries. Nonetheless, to use the correct reranking function, we must have access to the class probability distribution, which is inaccessible to black-box attacks, or some classical models such as 1-NN DTW, which uses distance-based computations to evaluate the nearest neighbor, may not even be able to calculate.\\
\hspace*{5mm}To address this limitiation, we use knowledge distillation a a method to train a student neural network $s$ that is equipped to replicate the predictions of the model to be targeted $f$. As such, we need to measure the attacked model's predictions on the dataset that we possess just one time, which can either be class labels or probability distributions across all classes. Then we use such labels as the ground truth labels that are conditioned to mimic the student $s$. We use one hot encoding scheme to measure the cross entropy loss in case the predictions are class marks, otherwise we try to imitate the distribution of probability directly. It should be remembered that the student model shares $D_{eval}$ with the GATN model of the training dataset.

\begin{figure*}[h]
\centering
\includegraphics[width=13cm]{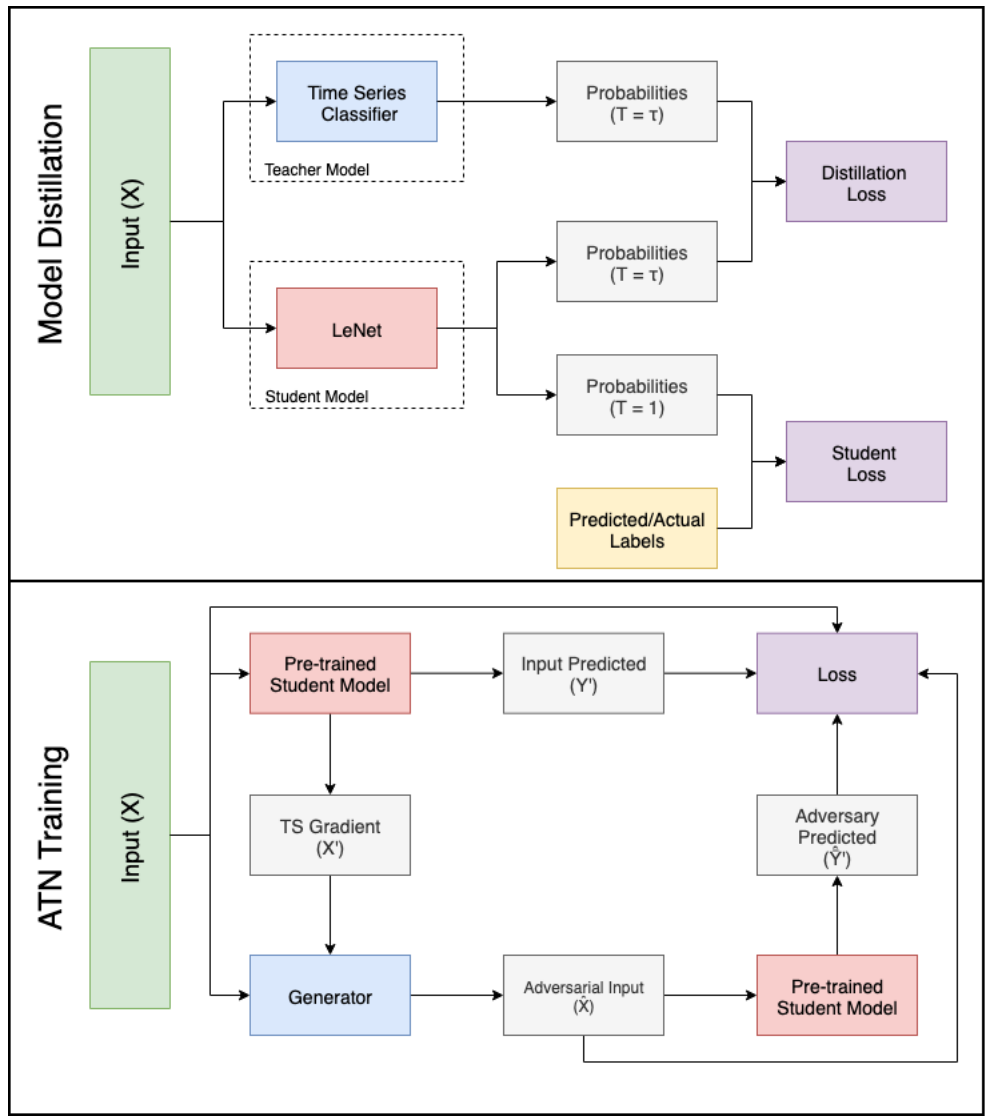}
\caption{The top diagram shows the methodology of training the model distillation used in the white-box and black-box attacks.
The bottom diagram is the methodology utilized to attack a time series classifier.}
\label{fig1}
\end{figure*}

\hspace*{5mm}As suggested by Hinton et al. \cite{hinton2015distilling}, we describe the training scheme of the student as shown in Figure \ref{fig1}. We scale the logits of the student $s$ and teacher $f$ (iff the teacher provides probabilities and it is a white-box attack) by a temperature scaling parameter $\tau$, which is kept constant at 10 for all experiments. When training the student model, we minimize the loss function defined as:
\begin{equation}
\label{eq:7}
    L_{transfer} = \gamma * L_{distillation} + (1-\gamma) * L_{student}
\end{equation}
\begin{equation}
\label{eq:8}
    L_{distillation} = \mathcal{H}(\sigma(z_f ; T= \tau), \sigma(z_s; T=\tau))
\end{equation}
\begin{equation}
\label{eq:9}
    L_{student} = \mathcal{H}(y, \sigma(z_s; T=1))
\end{equation}
where $\mathcal{H}$ is the standard cross entropy loss function, $z_s$ and $z_f$ are the un-normalized logits of the student ($s$) and teacher ($f$) models respectively, $\sigma(\cdot)$ is the scaled-softmax operation described in \eqref{eq:5}, $y$ is the ground truth labels, and $\gamma$ is a gating parameter between the two losses and is used to maintain a balance between how much the student $s$ imitates the teacher $f$ versus how much it learns from the hard label loss. When training a student as a white-box attack, we set $\gamma$ to be 0.5, allowing the equal weight to both losses, whereas for a black-box attack, we set $\gamma$ to be 1. Therefore for black-box attacks, we force the student $s$ to only mimic the teacher $f$ to the limit of its capacity. In setting this restriction, we limit the amount of information that may be made available to the GATN.

\subsection{Evaluation Methodology}
\label{section:3d}
\hspace*{5mm}We train the GATN on one of the two models because of the different restrictions between available information depending on whether the attack is a white-box or black-box attack. We assert that we train the GATN only when we conduct a white-box attack on a neural network by targeting the target neural network $f$ directly. In all other cases, whether the attack is a white-box or a black-box attack, and whether the attacked model is a neural network or a classical model, we pick the student model $s$ as the model attacked to train the GATN, and then use GATN's predictions ($\hat{x}$) to test whether the teacher model $f$ is also attacked if the expected adversarial input ($\hat{x}$) is used as a reference.\\
\hspace*{5mm}During the evaluation of the qualified GATN, we calculate the number of opponents of the $f$ model attacked that were obtained on the $D{ eval}$ training collection. We can calculate any metric in two situations during the assessment. Provided a split labelled dataset, we can double check whether or not an adversary has been detected. First, we check that the ground truth label matches the classifier's predicted label when supplied with an unmodified input ($y = y'$ when supplied with $x$ when supplied with $f$), and then check that this predicted label is different from the predicted label when supplied with the adversarial input ($y \neq \hat{y}'$ when supplied with $\hat{x}$ input). This ensures that we do not count an incorrect prediction from a random classifier as an attack.\\
\hspace*{5mm}Another circumstance is that we do not have any labeled samples prior to splitting the dataset. This training set is an unseen set for the attacked model $f$, therefore we consider that the dataset is unlabeled, and assume that the label predicted by the base classifier is the ground truth ($y = y'$ by default, when sample $x$ is provided to $f$). This is done prior to any attack by the GATN and is computed just once. We then define an adversarial sample as a sample $\hat{x}$ whose predicted class label is different than the predicted ground truth label ($y \neq \hat{y}$, when sample $\hat{x}$ is provided to $f$). A drawback of this approach is that it is overly optimistic and rewards sensitive classifiers that misclassify due to very minor alterations.
In order to adhere to an unbiased evaluation, we chose the first option, and use the labels we know from the test set to measure the adversarial inputs properly. In doing so, we consider the need for a labeled test set, but as shown above, following this method is not strictly necessary.

\section{Experiments and Results}
\label{section:4}
\hspace*{5mm}All methodologies were tested on 18 benchmark datasets for multivariate time series classification found in the UEA and UCR repository \cite{DBLP:journals/corr/abs-1811-00075}. Table \ref{table1} gives information about the multivariate time series. The evaluation has two objectives, to minimize the mean squared error (MSE) between the training dataset and the generated samples and; to maximize the number of adversaries for a set of chosen beta values. For all experiments, we keep $\alpha$, the reranking weight, set to 1.5, the target class set to 0, and perform a grid search over 5 possible values of $\beta$, the reconstruction weight term, such that $\beta = 10^{-b}$; $b \in \{1, 2, 3, 4, 5\}$. The code for all models are available at  \url{https://github.com/houshd/TS_Adv_multivariate}.

\begin{table*}[h]
\centering
 \caption{Dataset description for UEA and UCR multivariate time series benchmarks}
\label{table1}
\small

\begin{tabular}{|c|c|c|c|c|c|c|}
\hline
\textbf{Dataset} & \textbf{\begin{tabular}[c]{@{}c@{}}Train\\  Samples\end{tabular}}  & \textbf{\begin{tabular}[c]{@{}c@{}}Test\\  Samples\end{tabular}} & \textbf{Dimensions} & \textbf{\begin{tabular}[c]{@{}c@{}}Max Series\\  Length\end{tabular}} & \textbf{\begin{tabular}[c]{@{}c@{}}Num.\\  Classes\end{tabular}} \\ \hline
\textbf{ArticularyWordRecognition} & 275           & 300          & 9          & 144               & 25           \\ \hline
\textbf{AtrialFibrillation}        & 15            & 15           & 2          & 640               & 3             \\ \hline
\textbf{BasicMotions}              & 40            & 40           & 6          & 100               & 4            \\ \hline
\textbf{CharacterTrajectories}     & 1422          & 1436         & 3          & 182               & 20           \\ \hline
\textbf{Cricket}                   & 108           & 72           & 6          & 1197              & 12           \\ \hline
\textbf{Epilepsy}                  & 137           & 138          & 3          & 206               & 4             \\ \hline
\textbf{EthanolConcentration}      & 261           & 263          & 3          & 1751              & 4             \\ \hline
\textbf{ERing}                     & 30            & 270          & 4          & 65                & 6             \\ \hline
\textbf{FingerMovements}           & 316           & 100          & 28         & 50                & 2            \\ \hline
\textbf{HandMovementDirection}     & 160           & 74           & 10         & 400               & 4             \\ \hline
\textbf{Handwriting}               & 150           & 850          & 3          & 152               & 26            \\ \hline
\textbf{JapeneseVowels}            & 270           & 370          & 12         & 29                & 9             \\ \hline
\textbf{Libras}                    & 180           & 180          & 2          & 45                & 15           \\ \hline
\textbf{LSST}                      & 2459          & 2466         & 6          & 36                & 14            \\ \hline
\textbf{NATOPS}                    & 180           & 180          & 24         & 51                & 6             \\ \hline
\textbf{PenDigits}                 & 7494          & 3498         & 2          & 8                 & 10            \\ \hline
\textbf{RacketSports}              & 151           & 152          & 6          & 30                & 4             \\ \hline
\textbf{UWaveGestureLibrary}       & 120           & 320          & 3          & 315               & 8             \\ \hline
\end{tabular}

\end{table*}

\subsection{Experiments}
\hspace*{5mm}In this study, both neural networks and traditional time series classifiers were chosen as the model $f$ to target. We use a Fully Convolutional Network for the attacked neural network and 1-NN Dynamic Time Warping is used for the traditional base model.\\
\hspace*{5mm}To retain the strictest definition of black and white-box attacks, we only use the attacked model's discrete class label for black-box attacks and use the probability distribution expected by the white-box attack classifier. The only exception where a student-teacher network is not used is when conducting a white-box attack on a FCN time series model, since an Adversarial Transformation Network (ATN) can directly manipulate the gradient information from a neural network. The performance of the adversarial model is evaluated on the teacher classification model for the original time series.\\
\hspace*{5mm}For every student model we train, we utilize the LeNet-5 architecture \cite{lecun2015lenet}. The LeNet-5 time series classifier is defined as a classical Convolutional Neural network following the structure: Conv (6 filters, 5x5, valid padding) -> Max Pooling -> Conv (16 filters, 5x5, valid padding) -> Max Pooling -> Fully Connected (120 units, relu) -> Fully Connected (84 units, relu) -> Fully Connected (number of classes, softmax).\\
\hspace*{5mm}The fully convolutional network is an exptension of the FCN model proposed by Wang et al \cite{wang2017time}. The FCN is comprised of three blocks, each comprised of a sequence of Convolution layer -> Batch Normalization -> ReLU activations. All convolutional kernels are initialized using the uniform he initialization proposed by He et al \cite{he2015delving}. We utilize filters of size [128, 256, 128] and kernel of size of [8, 5, 3].\\
\hspace*{5mm}One strong baseline deterministic model for classifying multivariate time series is 1-NN DTW without a warping window. The distance based nature of the 1-NN classifier and the reliance on a distance matrix, 1-NN DTW cannot easily be used to compute an equivalent soft probabilistic representation. Since white-box attacks have access to only the probability distribution predicted for each sample, the distance matrix generated by DTW is used to compute an equivalent soft probabilistic representation. The analogous representation is such that we get the exact same result as selecting the 1-NN on the real distance matrix if we determine the top class on this representation\\
\hspace*{5mm}To compute this soft probabilistic representation, consider a set of distance matrices $V$ computed using a distance measure such as DTW between all possible pairs of samples between the two datasets being compared.

\begin{algorithm}[h]
 \caption{Equivalent Probabilistic Representation of the
Distance Matrix for 1-Nearest Neighbor Classification}
 \label{alg:1}
 \KwData{$V$ is a distance matrix of shape $[N_{test}; N_{train}; N_{Channels}]$ and $y$ is the train set label vector of length $N_{train}$}
 \KwResult{Softmax normalized predictions $p$ of shape $[N_{test};C]$ and the discrete label vector $q$ of length $N_{test}$}
    \Begin{
    $V \longleftarrow (-V)$\\
    $uniqueLabels = Unique(y)$ //\small{Unique Class Labels} \\
    $V_c = []$\\

    \For{$c_i \in uniqueLabels$}{
        $v_c = V_{(y=c_i)}$ //\small{$[N_{test}; N_{train}(y = c_i); N_{Channels}]$}\\
        $v_c\_max = max(v_c)$ //\small{$[N_{test}]$}\\
        $V_c.append(v_c\_max)$\\
    }
    $V' = concatenate(V_c)$ //\small{$[N_{test};$ \# of classes$]$}\\
    $p = softmax(V')$ //\small{$[N_{test};$ \# of classes$]$}\\
    $q = argmax(p)$ //\small{$[N_{test}]$}\\
    
    \textbf{return} (p,q)
    }

\end{algorithm}

Algorithm \ref{alg:1} is an intermediate standardization algorithm that accepts a set of $V$ distance matrices and the training class labels of $y$ as inputs, and calculates an analogous probabilistic representation that can be used directly to determine the 1-Nearest Neighbor. The Soft-1NN algorithm selects all samples that belong to a class $c_i$, where $i \in \{1, \dots , C\}$ as $v_c$, computes the maximum over all train samples for that class, then appends the vector $v_c\_max$ to the list $V_c$. The concatenation of all of these lists of vectors in $V_c$ then represents the matrix $V'$, which is passed to the softmax function, as shown in Equation \ref{eq:5} with $T$ set to 1, to represent this matrix $V'$ as a probabilistic equivalent of the original distance matrix $V$. An implicit restriction placed on Algorithm \ref{alg:1} is that the representation is equivalent only when computing the 1-NN DTW. It cannot be used to to represent the K-NN DTW (or any other distance metric) and therefore cannot be used for K-NN classification. However, in time series classification, the value of K is typically set to 1 for nearest neighbor classifiers. While the Algorithm \ref{alg:1} has been used to convert the 1-NN DTW distance matrices, any set of distance matrices used for 1-NN classification algorithms can also be used to standardize it.

\subsection{Results}
\begin{figure*}[]
\centering
\includegraphics[width=13cm]{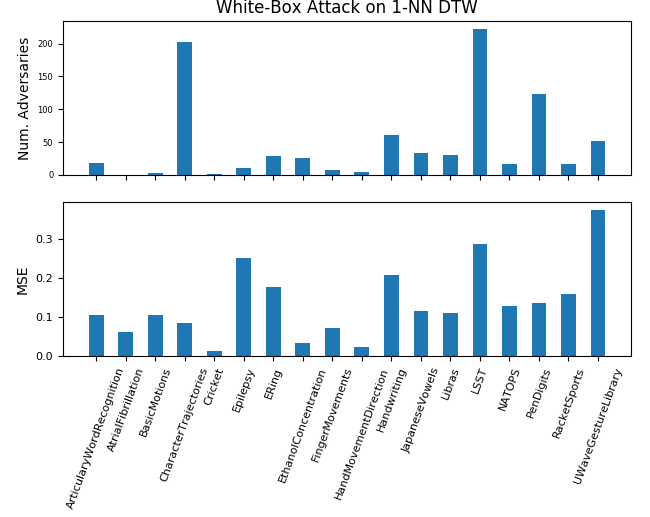}
\caption{White-box attack on 1-NN DTW that is trained on 18 datasets}
\label{wb_dtw}
\end{figure*}

\begin{figure*}[]
\centering
\includegraphics[width=13cm]{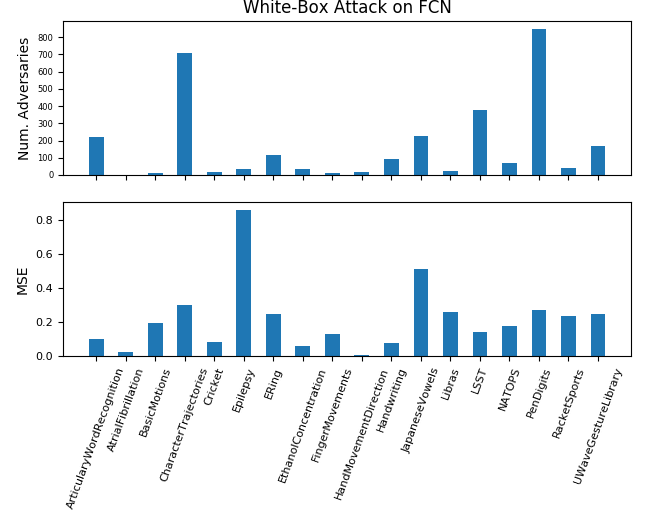}
\caption{White-box attack on FCN that is trained on 18 datasets}
\label{wb_nn}
\end{figure*}

\begin{figure*}[]
\centering
\includegraphics[width=13cm]{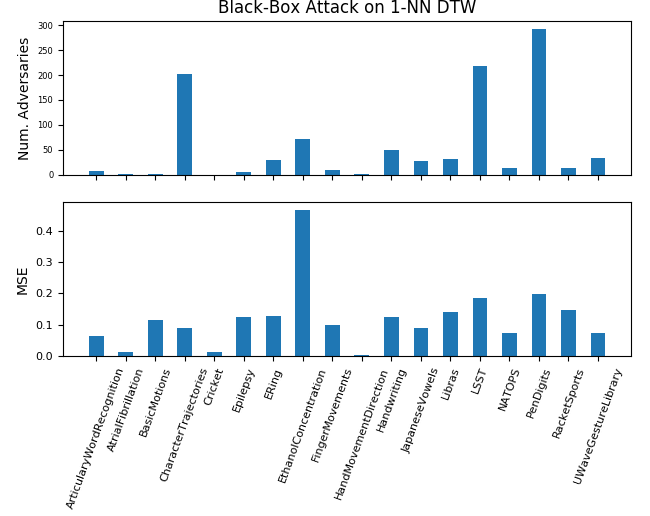}
\caption{Black-box attack on 1-NN DTW that is trained on 18 datasets}
\label{bb_dtw}
\end{figure*}

\begin{figure*}[]
\centering
\includegraphics[width=13cm]{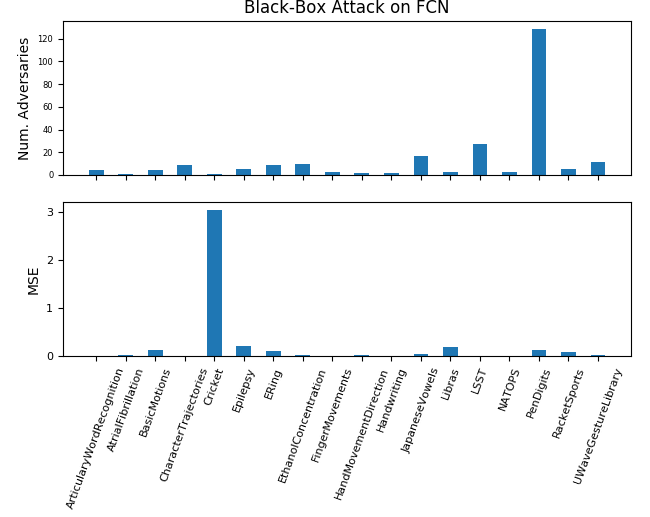}
\caption{Black-box attack on FCN that is trained on 18 datasets}
\label{bb_nn}
\end{figure*}

\begin{figure*}[]
\centering
\includegraphics[width=13cm]{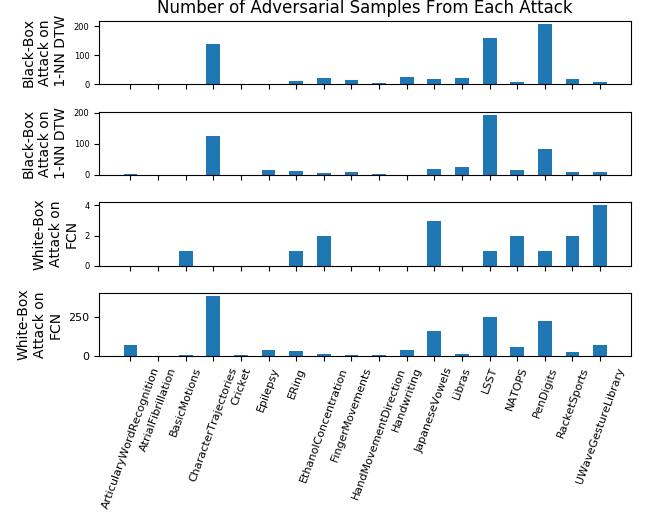}
\caption{Black-box and white-box attacks on FCN and 1-NN DTW classifiers that are tested on $D_{test}$ without any retraining.}
\label{testfig}
\end{figure*}

\hspace*{5mm}Figures \ref{wb_dtw} and \ref{wb_nn} depict the results from white-box attacks on 1-NN DTW and FCN that are applied on 18 multivariate time series datasets. Figures \ref{bb_dtw} and \ref{bb_nn} represent the results from black-box attacks on 1-NN DTW and FCN classifiers trained on the same 18 datasets. Experimental results for these 18 dataset aimed to generate adversaries for only one class. The detailed results can be found in Appendix A. The proposed methodology is successful in capturing adversaries on all datasets. \\
\hspace*{5mm}The number of adversaries in each dataset and the amount of perturbation per sample are reliant on the hyper-parameters being tested on. As an example, the dataset “AtrialFibrillation” was only able to generate multiple adversaries for the black-box attack on 1-NN DTW when the target class was 0. However, if the target class is 1, the number of adversaries generated increases to 5, 2, 3, 13 for a black-box attack on 1-NN DTW, white-box attack on 1-NN DTW, black-box attack on FCN and white-box attack on FCN, respectively. These numbers could potentially be higher if the hyper-parameters are optimized for this target class. Additionally, the target class has a significant impact on the adversary being produced because of the ATN's loss function. For time series groups, it is easier to generate adversaries which are identical to one another.
\hspace*{5mm}A Wilcoxson signed-rank test is utilized to compare the number of adversaries generated by white-box and black-box attacks on FCN and 1-NN DTW classifiers that are trained on the 18 datasets, summarized in Table \ref{table2}. Our findings indicate that the FCN classifier is more susceptible to a white-box attack compared with a 1-NN DTW white-box attack. It should be noted that the FCN classifier white-box attack is producing considerably more adversaries than its counterparts. This is because the attack with the white-box is explicitly on the FCN model and not on a student model approximating the classifier's behaviour. We observe that the number of adversarial samples obtained on FCN classifiers from black-box attacks is greater than the number of adversarial samples from either white-box or black-box attacks on DTW classifiers. A Wilcoxson signed-rank test supports this finding by showing a statistically significant difference (at a rate of 0.05) in the number of adversarial samples observed on 1-NN DTW classifiers due to black-box or white-box attacks versus the number of adversarial samples obtained by black-box attacks on FCN classifiers. We also detect that 1-NN DTW classifiers under either type of attack have approximately the same number of adversaries generated. Finally, we find that FCNs has the least number of adversarial samples after black-box attacks, although each of these samples requires indistinguishable disturbances to the original signal. These observations are important to future research into the development of time series classifiers, as the number of adversarial samples generated under each methodology can be used as a secondary evaluation metric to measure the robustness of a model. The average MSE of adversarial samples after black-box attacks on FCN classifiers is lower than the average MSE of the adversarial samples obtained via black-box and white-box attacks on 1-NN DTW classifiers, but only statistically significant when compared to white-box, as observed in Table \ref{table3}. A lower MSE indicates the black-box attack on FCN classifiers requires minimal perturbations per time series sample in comparison to the attacks on 1-NN DTW classifiers. \\
\hspace*{5mm}Now we test how well GATN generalizes onto an unseen dataset, $D_{test}$, such that GATN does not require any additional training. This is beneficial in situations where the time series adversarial samples are generated in constant time of a single forward pass of the GATN model without requiring further training. Such a generalization is uncommon to adversarial methodologies because they require retraining to generate adversarial samples. Our proposed methodology is robust, successfully generating adversarial samples on data that is unseen to both the GATN and the student models, for the respective targeted time series classification models. Figure \ref{testfig} depicts the number of adversarial samples detected, on an unseen dataset, with a white-box and black-box attack on the 1-NN DTW classifiers and FCN classifiers. The white-box attack on the FCN classifier obtains the most adversarial samples per dataset. This is followed by a white-box and black-box attack on the 1-NN DTW, which show similar number of adversarial samples constructed. Finally, we find that the FCN classifier is the least susceptible to black-box attacks. \\
\hspace*{5mm}The unique consequence of this generalization is the application of trained GATN models for attacks that are feasible on real world devices, even for black-box attacks. The deployment of a trained GATN with the paired student model affords a near constant-time cost of generating reasonable number of adversarial samples. As the forward pass of the GATN requires few resources, and the student model is small enough to compute the input gradient ($\tilde{x}$) in reasonable time, these attacks can be constructed without significant computation on small, portable devices. Therefore, the fact that certain classifiers that are trained on certain datasets can be attacked without requiring any additional on-device training is concerning.

\begin{table}[h]
 \caption{Wilcoxson signed-rank test comparing the number of adversaries between the different attacks}
\label{table2}
\center
\begin{tabular}{|c|c|c|c|}
\hline
                                                                       & \textbf{\begin{tabular}[c]{@{}c@{}}White-box \\ 1-NN DTW\end{tabular}} & \textbf{\begin{tabular}[c]{@{}c@{}}Black-box \\ FCN\end{tabular}} & \textbf{\begin{tabular}[c]{@{}c@{}}White-box \\ FCN\end{tabular}} \\ \hline
\textbf{\begin{tabular}[c]{@{}c@{}}Black-box \\ 1-NN DTW\end{tabular}} & \cellcolor[HTML]{FFFFFF}{\color[HTML]{333333} 1.99E-01}                & \cellcolor[HTML]{67FD9A}{\color[HTML]{036400} 1.02E-03}           & \cellcolor[HTML]{67FD9A}{\color[HTML]{036400} 1.59E-03}           \\ \hline
\textbf{\begin{tabular}[c]{@{}c@{}}White-box \\ 1-NN DTW\end{tabular}} & \cellcolor[HTML]{333333}{\color[HTML]{333333} }                        & \cellcolor[HTML]{67FD9A}{\color[HTML]{036400} 1.28E-03}           & \cellcolor[HTML]{67FD9A}{\color[HTML]{036400} 3.26E-04}           \\ \hline
\textbf{\begin{tabular}[c]{@{}c@{}}Black-box \\ FCN\end{tabular}}      & \cellcolor[HTML]{333333}                                               & \cellcolor[HTML]{333333}                                          & \cellcolor[HTML]{67FD9A}{\color[HTML]{036400} 2.92E-04}           \\ \hline
\end{tabular}
\end{table}

\begin{table}[h]
 \caption{Wilcoxson signed-rank test comparing the MSE between the different attacks}
\label{table3}
\center

\begin{tabular}{|c|c|c|c|}
\hline
                                                                       & \textbf{\begin{tabular}[c]{@{}c@{}}White-box \\ 1-NN DTW\end{tabular}} & \textbf{\begin{tabular}[c]{@{}c@{}}Black-box \\ FCN\end{tabular}} & \textbf{\begin{tabular}[c]{@{}c@{}}White-box \\ FCN\end{tabular}} \\ \hline
\textbf{\begin{tabular}[c]{@{}c@{}}Black-box \\ 1-NN DTW\end{tabular}} & \cellcolor[HTML]{FFFFFF}{\color[HTML]{333333} 1.33E-01}                & \cellcolor[HTML]{FFFFFF}{\color[HTML]{333333} 6.42E-02}           & \cellcolor[HTML]{67FD9A}{\color[HTML]{036400} 1.08E-02}           \\ \hline
\textbf{\begin{tabular}[c]{@{}c@{}}White-box \\ 1-NN DTW\end{tabular}} & \cellcolor[HTML]{333333}{\color[HTML]{333333} }                        & \cellcolor[HTML]{67FD9A}{\color[HTML]{036400} 2.49E-02}           & \cellcolor[HTML]{FFFFFF}{\color[HTML]{333333} 7.07E-02}           \\ \hline
\textbf{\begin{tabular}[c]{@{}c@{}}Black-box \\ FCN\end{tabular}}      & \cellcolor[HTML]{333333}                                               & \cellcolor[HTML]{333333}                                          & \cellcolor[HTML]{67FD9A}{\color[HTML]{036400} 4.34E-03}           \\ \hline
\end{tabular}
\end{table}

\section{Conclusion}
\label{section:5}
\hspace*{5mm}This work proposes a model distillation technique to mimic the behavior of the various classical multivariate time series classification models and an adversarial transformation network to attack various multivariate time series datasets. The proposed methodology is applied onto 1-NN DTW and Fully Connected Network (FCN) that are trained on 18 University of East Anglia (UEA) and University of California Riverside (UCR) datasets. All 18 datasets showed to be susceptible to at least one type of attack. To the best of our knowledge, this is the first time multivariate time series classifiers have been attacked using adversarial neural networks. The classical 1-NN DTW proved to be more robust to adversarial attacks than the FCN model. We suggest potential researchers creating models for the classification of time series to take model robustness as an evaluative metric. In addition, we suggest that adversarial data samples be integrated into their training data sets to further enhance resistance to adversarial attacks.

\FloatBarrier

\appendices
\section{Detailed Results}

\begin{table}[h]
 \caption{Black-box attack on 1-NN DTW models}
\label{table4}
\center
\begin{tabular}{|c|c|c|}
\hline
\textbf{Dataset}             & \textbf{Num. of Adversaries} & \textbf{MSE} \\ \hline
ArticularyWordRecognition & 8                            & 0.063085     \\ \hline
AtrialFibrillation        & 2                            & 0.015005     \\ \hline
BasicMotions              & 2                            & 0.114111     \\ \hline
CharacterTrajectories     & 202                          & 0.08986      \\ \hline
Cricket                   & 0                            & 0.012914     \\ \hline
Epilepsy                  & 5                            & 0.123488     \\ \hline
ERing                     & 30                           & 0.128423     \\ \hline
EthanolConcentration      & 72                           & 0.466593     \\ \hline
FingerMovements           & 10                           & 0.09904      \\ \hline
HandMovementDirection     & 2                            & 0.004315     \\ \hline
Handwriting               & 49                           & 0.125444     \\ \hline
JapaneseVowels            & 28                           & 0.090313     \\ \hline
Libras                    & 31                           & 0.142043     \\ \hline
LSST                      & 219                          & 0.185534     \\ \hline
NATOPS                    & 14                           & 0.073556     \\ \hline
PenDigits                 & 293                          & 0.197882     \\ \hline
RacketSports              & 14                           & 0.147109     \\ \hline
UWaveGestureLibrary       & 34                           & 0.075413     \\ \hline
\end{tabular}
\end{table}

\begin{table}[h]
 \caption{White-box attack on 1-NN DTW models}
\label{table5}
\center
\begin{tabular}{|c|c|c|}
\hline
\textbf{Dataset}             & \textbf{Num. of Adversaries} & \textbf{MSE} \\ \hline
ArticularyWordRecognition & 18                           & 0.104909     \\ \hline
AtrialFibrillation        & 0                            & 0.061174     \\ \hline
BasicMotions              & 3                            & 0.105032     \\ \hline
CharacterTrajectories     & 203                          & 0.084731     \\ \hline
Cricket                   & 1                            & 0.013019     \\ \hline
Epilepsy                  & 11                           & 0.249629     \\ \hline
ERing                     & 29                           & 0.176111     \\ \hline
EthanolConcentration      & 26                           & 0.033794     \\ \hline
FingerMovements           & 8                            & 0.071567     \\ \hline
HandMovementDirection     & 4                            & 0.023995     \\ \hline
Handwriting               & 61                           & 0.206134     \\ \hline
JapaneseVowels            & 33                           & 0.114339     \\ \hline
Libras                    & 31                           & 0.109957     \\ \hline
LSST                      & 223                          & 0.285733     \\ \hline
NATOPS                    & 16                           & 0.129278     \\ \hline
PenDigits                 & 123                          & 0.136015     \\ \hline
RacketSports              & 16                           & 0.157508     \\ \hline
UWaveGestureLibrary       & 52                           & 0.373157     \\ \hline
\end{tabular}
\end{table}

\begin{table}[h]
 \caption{Black-box attack on FCN models}
\label{table6}
\center
\begin{tabular}{|c|c|c|}
\hline
\textbf{Dataset}             & \textbf{Num. of Adversaries} & \textbf{MSE} \\ \hline
ArticularyWordRecognition & 4                            & 0.000156     \\ \hline
AtrialFibrillation        & 1                            & 0.016261     \\ \hline
BasicMotions              & 4                            & 0.131301     \\ \hline
CharacterTrajectories     & 9                            & 0.002883     \\ \hline
Cricket                   & 1                            & 3.046668     \\ \hline
Epilepsy                  & 5                            & 0.203446     \\ \hline
ERing                     & 9                            & 0.10876      \\ \hline
EthanolConcentration      & 10                           & 0.019332     \\ \hline
FingerMovements           & 3                            & 0.0031       \\ \hline
HandMovementDirection     & 2                            & 0.018381     \\ \hline
Handwriting               & 2                            & 0.012477     \\ \hline
JapaneseVowels            & 17                           & 0.043202     \\ \hline
Libras                    & 3                            & 0.200899     \\ \hline
LSST                      & 27                           & 0.01518      \\ \hline
NATOPS                    & 3                            & 0.011849     \\ \hline
PenDigits                 & 129                          & 0.129774     \\ \hline
RacketSports              & 5                            & 0.096207     \\ \hline
UWaveGestureLibrary       & 11                           & 0.016174     \\ \hline
\end{tabular}
\end{table}

\begin{table}[h]
 \caption{White-box attack on FCN models}
\label{table7}
\center
\begin{tabular}{|c|c|c|}
\hline
\textbf{Dataset}             & \textbf{Num. of Adversaries} & \textbf{MSE} \\ \hline
ArticularyWordRecognition & 223                          & 0.103408     \\ \hline
AtrialFibrillation        & 1                            & 0.027168     \\ \hline
BasicMotions              & 14                           & 0.196938     \\ \hline
CharacterTrajectories     & 706                          & 0.300598     \\ \hline
Cricket                   & 15                           & 0.083996     \\ \hline
Epilepsy                  & 32                           & 0.859604     \\ \hline
ERing                     & 118                          & 0.250408     \\ \hline
EthanolConcentration      & 37                           & 0.05864      \\ \hline
FingerMovements           & 13                           & 0.127995     \\ \hline
HandMovementDirection     & 17                           & 0.006507     \\ \hline
Handwriting               & 94                           & 0.076364     \\ \hline
JapaneseVowels            & 224                          & 0.512852     \\ \hline
Libras                    & 24                           & 0.259869     \\ \hline
LSST                      & 376                          & 0.141442     \\ \hline
NATOPS                    & 69                           & 0.178026     \\ \hline
PenDigits                 & 850                          & 0.270337     \\ \hline
RacketSports              & 39                           & 0.236502     \\ \hline
UWaveGestureLibrary       & 166                          & 0.248728     \\ \hline
\end{tabular}
\end{table}

\section*{Acknowledgment}
The authors would like to thank all the researchers that helped create and clean the data available in the updated UEA and UCR Multivariate Time Series Classification Archive. Sustained research in this domain would be much more challenging without their efforts.

\FloatBarrier
\bibliographystyle{unsrt}  
\bibliography{biblio}  %%% Remove comment to use the external .bib file (using bibtex).

\begin{IEEEbiography}[{\includegraphics[width=1in,height=1.5in,clip,keepaspectratio]{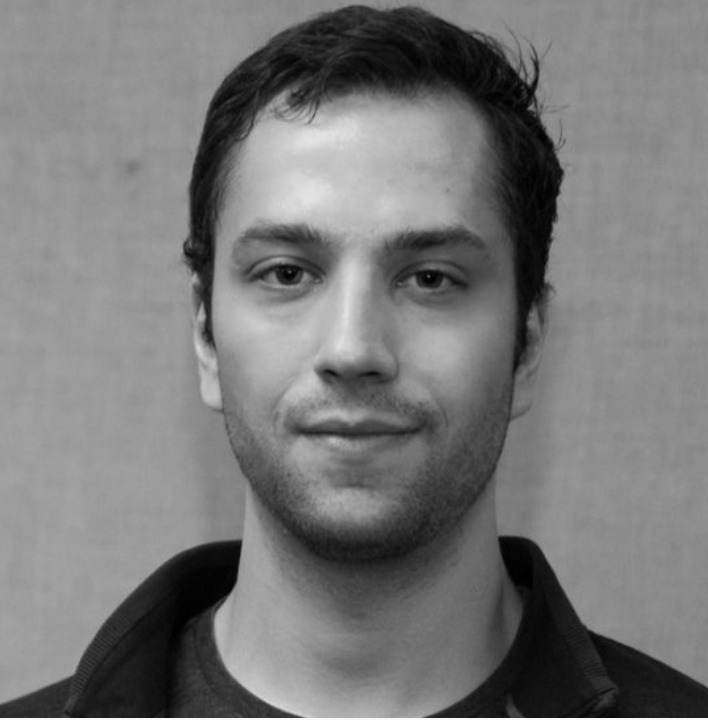}}]{Samuel Harford} received the B.S. degree in Industrial Engineering from the University of Illinois at Chicago in 2016, the M.S. degree in Industrial Engineering from the University of Illinois at Chicago in 2018. He is currently pursuing the Ph.D. degree with the Mechanical and Industrial Engineering Department, University of Illinois at Chicago. His research interests lie in the domain of machine learning, time series classification, and heath care data mining.
\end{IEEEbiography}

\begin{IEEEbiography}[{\includegraphics[width=1in,height=1.5in,clip,keepaspectratio]{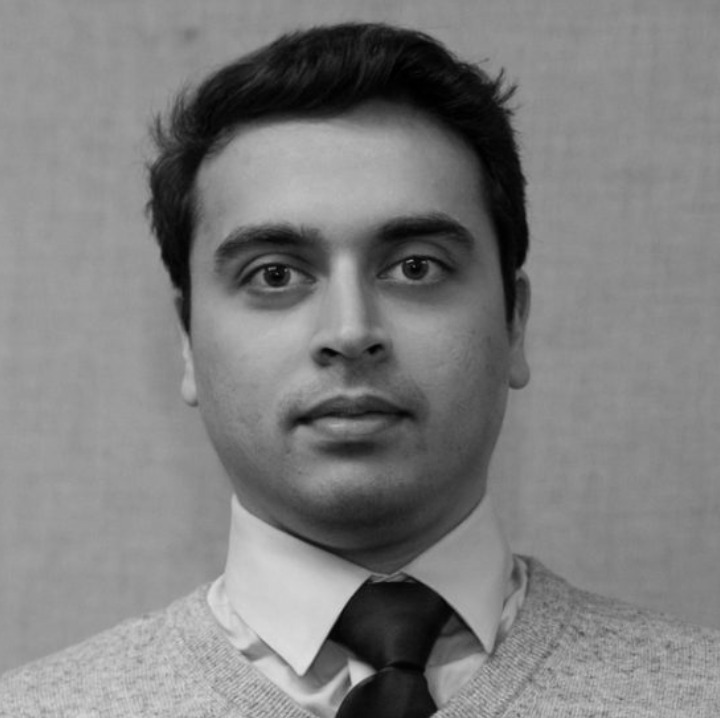}}]{Fazle Karim} received the B.Sc. degree in industrial engineering from the University of Illinois at Urbana-Champaign in 2012, the M.Sc. degree in industrial engineering from the University of Illinois at Chicago in 2016. He is currently pursuing the Ph.D. degree with the Mechanical and Industrial Engineering Department, University of Illinois at Chicago. He is also the Lead Data Scientist with Prominent Laboratory, the university’s foremost research facility in process mining. His research interests include education data mining, health care data mining, and time series analysis.
\end{IEEEbiography}

\begin{IEEEbiography}[{\includegraphics[width=1in,height=1.5in,clip,keepaspectratio]{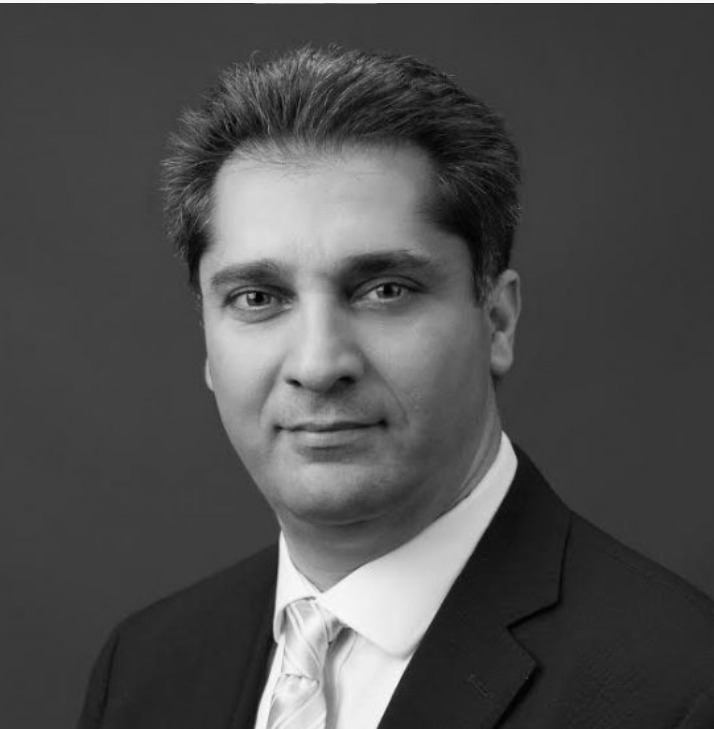}}]{Houshang Darabi}(S'98-A'00-M'10-SM'14) received the Ph.D. degree in Industrial and Systems Engineering from Rutgers University, New Brunswick, NJ, USA, in 2000.\\ He is currently an Associate Professor with the Department of Mechanical and Industrial Engineering, University of Illinois at Chicago (UIC), and also an Associate Professor with the Department of Computer Science, UIC. He has been a contributing author of two books in the areas of scalable enterprise systems and reconfigurable discrete event systems. His research has been supported by several federal and private agencies, such as the National Science Foundation, the National Institute of Standard and Technology, the Department of Energy, and Motorola. He has extensively published on various automation and project management subjects, including wireless sensory networks for location sensing, planning and management of projects with tasks requiring multi-mode resources, and workflow modeling and management. He has published in different prestigious journals and conference proceedings, such as the IEEE Transaction on Robotics and Automation, the IEEE Transactions on Automation Science and Engineering, and the IEEE Transactions on Systems, Man, and Cybernetics, and Information Sciences. His current research interests include the application of data mining, process mining, and optimization in design and analysis of manufacturing, business, project management, and workflow management systems.\end{IEEEbiography}

\end{document}